\documentclass[10pt, conference, letterpaper]{IEEEtran}

\usepackage[english]{babel}
\usepackage{blindtext}
\usepackage{graphicx}
\usepackage{epstopdf}
\usepackage{amsfonts,amsmath,color,amssymb,amsxtra,amsbsy,floatflt}
\usepackage{amsthm}
\usepackage{textcomp}       
\usepackage{indentfirst} 
\usepackage{url}
\usepackage{color}
\usepackage{marvosym}
\usepackage{algorithm}
\usepackage{algpseudocode}
\usepackage{comment}
\usepackage{caption}
\usepackage{tabularx}
\usepackage{hhline}
\usepackage{cite}
\usepackage{textcomp}
\usepackage{multirow}
\usepackage[table]{xcolor}      

\usepackage{subcaption}

\graphicspath{ {Figures/} }

\newcommand{\name}{$\pi$-ROAD}
\newtheorem{problem}{Problem}
\newtheorem{proposition}{Proposition}

\usepackage{flushend}
\renewcommand{\baselinestretch}{0.985}

\begin{document}

\title{\name{}: a Learn-as-You-Go Framework for\\ On-Demand Emergency Slices in V2X Scenarios
}

\author{\IEEEauthorblockN{Armin~Okic\IEEEauthorrefmark{1},
Lanfranco~Zanzi\IEEEauthorrefmark{2}\IEEEauthorrefmark{3}, Vincenzo~Sciancalepore\IEEEauthorrefmark{2},
Alessandro~Redondi\IEEEauthorrefmark{1}, and Xavier~Costa-P\'erez\IEEEauthorrefmark{2}\IEEEauthorrefmark{4}}
\IEEEauthorblockA{
\IEEEauthorrefmark{1}Politecnico di Milano, Milano, Italy, Email:\{name.surname\}@polimi.it}
\IEEEauthorrefmark{2}NEC Laboratories Europe, Heidelberg, Germany,
Email:\{name.surname\}@neclab.eu,\\
\IEEEauthorrefmark{3} Technische Universit\"at Kaiserslautern, Kaiserslautern, Germany\\
\IEEEauthorrefmark{4} i2CAT Foundation and ICREA, Barcelona, Spain.
\\
}

\maketitle

\begin{abstract}

Vehicle-to-everything (V2X) is expected to become one of the main drivers of 5G business in the near future.
Dedicated \emph{network slices} are envisioned to satisfy the stringent requirements of advanced V2X services, such as autonomous driving, aimed at drastically reducing road casualties. 
However, as V2X services become more mission-critical, new solutions need to be devised to guarantee their successful service delivery even in exceptional situations, e.g. road accidents, congestion, etc. 
In this context, we propose \name{}, a \emph{deep learning} framework to automatically learn regular mobile traffic patterns along roads, detect non-recurring events and classify them by severity level. \name{} enables operators to \emph{proactively} instantiate dedicated \emph{Emergency Network Slices (ENS)} as needed while re-dimensioning the existing slices according to their service criticality level.
Our framework is validated by means of real mobile network traces collected within $400~km$ of a highway in Europe and augmented with publicly available information on related road events. Our results show that \name{} successfully detects and classifies non-recurring road events and reduces up to $30\%$ the impact of ENS on already running services.

\end{abstract}

\section{Introduction}
\label{sect:introduction}

Despite the high investment volume in public transportation, many people daily commute to work with their private vehicles over major roads around cities. 
Accordingly, both drivers and passengers consume and generate a large amount of data along the road mobile infrastructure for a wide variety of purposes: navigation systems, car sensors, infotainment but also phone calls, social media, streaming, etc.
Along highly crowded roads, this may lead to network congestion and, in the worst case, service disruptions.

In order to avoid these situations, the next generation of mobile networks (5G) has defined the \emph{Network Slicing} concept, 
which allows infrastructure providers to dynamically instantiate \emph{on-demand customized virtual network instances} with dedicated Service Level Agreements (SLAs). Standardization bodies have defined the overarching architecture~\cite{TS23.501} to support such isolated slices, thus fostering research on dynamic resource orchestration mechanisms based on well-known technologies such as Network Function Virtualization (NFV) and Software Defined Networking (SDN)~\cite{Network_Slicing_to_Enable_Scalability,Network_Slicing_with_SDN_NFV}.

As 5G gets rolled-out and advanced V2X services deployed, solutions to protect mission-critical services will become increasingly important. While significant road safety improvements have been introduced in the last decades, road fatalities are still a major cause of injuries and death worldwide ~\cite{WHO_Accidents_Report}.
Thus, guaranteeing public safety along roads is still a major challenge that requires novel solutions. Due to the combined effect of high-mobility patterns, traffic volumes and advanced automotive-related use cases (V2X), mobile network infrastructure along roads will face daunting challenges to guarantee mission critical services in unexpected congestion scenarios.
In this context, \emph{Emergency Network Slices (ENS)} are expected not only to improve situation awareness during emergencies but also to support the provisioning of enhanced communication schemes, e.g., Ultra-Reliable and Low-latency Communication (URLLC) for virtual and augmented reality (VR/AR) to first-responder teams, e.g., ambulances, police, firefighters, that have to reach the event location and manage emergencies in a faster and safer manner~\cite{Emergency_Vehicles_crashes_report}. As ENS will get top priority, solutions need to be devised to re-dimension already running services according to their criticality level. 
Unexpected road events progressively cause traffic congestion to nearby areas and, in turn, saturate the networking resources of adjacent base stations. If such a \emph{propagation effect} could be \emph{predicted}, their congestion effects could be alleviated by means of \emph{proactive slices resource management}.

We take on this challenge and propose \emph{Passive Information-based Resource Orchestration in Automotive Driving scenarios (\name{})} that relies on a deep learning framework to \emph{analyze} passive information from real-time mobile network traffic statistics, \emph{learn} regular traffic patterns, and accurately \emph{detect} anomalous deviations due to unexpected road events.

Our contributions can be summarized as follows: $i$) we provide an in-depth analysis of \emph{real mobile network traces} and statistics from an operational network along 400 Km of a major road in Europe, $ii$) we design a deep learning model, namely \name{}, that applies the well-known concept of learn-as-you-go, i.e., it accurately detects and localizes road events and classifies them according to their severity, $iii$) we focus on the network orchestration of emergency scenarios where an Emergency Network Slice (ENS) must be instantiated according to the type of road event, $iv$) we formulate an optimization problem to minimize the impact of ENSs on already existing slices and $v$) we validate our model and assess its capabilities by means of a full dataset of network monitoring data in realistic scenarios.

The remainder of the paper is structured as follows. Section~\ref{sect:data_analysis} provides an overview of road dynamics from a mobile network perspective. Section~\ref{sect:design} details the main building blocks of our solution. Leveraging on the output of \name{}, Section~\ref{sect:framework} formulates the ENS problem, whereas Section~\ref{sect:perf_eval} presents an exhaustive simulation campaign to validate our design principles through real operational data. Section~\ref{sect:related} summarizes the related literature and \ref{sect:conclusions} concludes the paper.

\paragraph*{Privacy issues}
The research activity presented in this paper does not violate user's privacy rights. The dataset contains only aggregated and anonymous information collected by the network operator for monitoring purposes.

\section{Data Analysis}
\label{sect:data_analysis}

We carry out an analysis of the network traffic dynamics occurring on the vehicular roads.

\begin{figure}[t!]
\centering
\includegraphics[width=\columnwidth, clip, trim = 3.5cm 2.5cm 2cm 2cm]{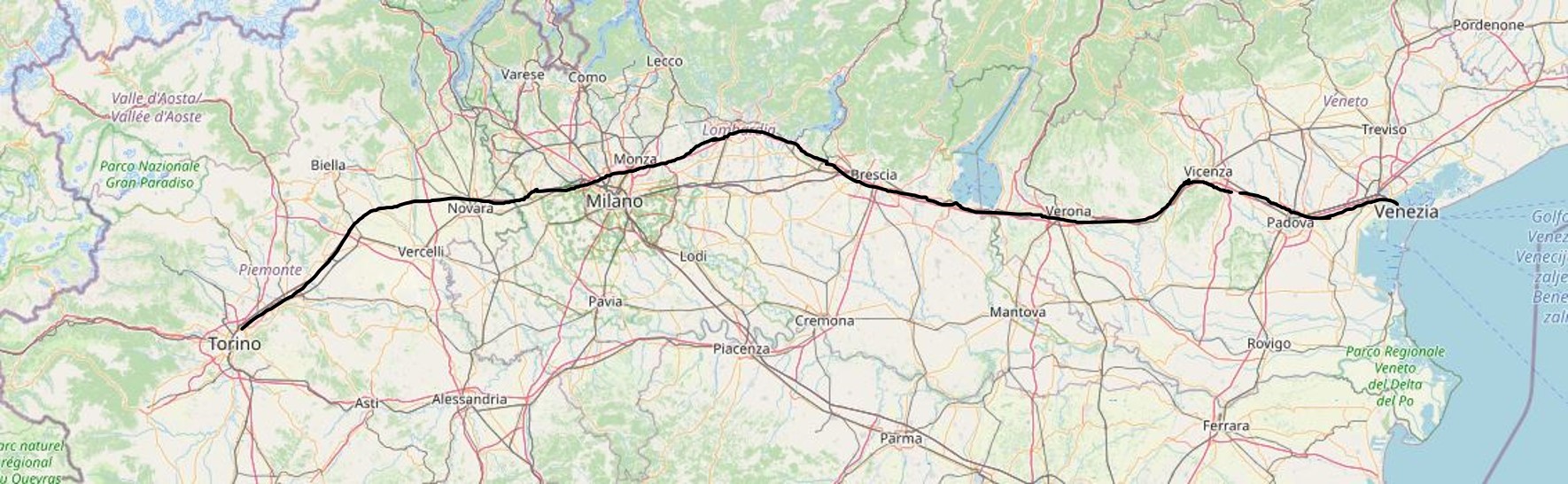}
\caption{\small Overview of the Italian highway (A4).}
\label{fig:highway_map}
\end{figure}

The considered dataset consists of $6$ months of real network data (February-July $2020$) collected from an operational LTE network deployed alongside the Italian A4 highway shown in Fig.~\ref{fig:highway_map}. 
The highway has a length of approximately $400$ km and is located in the north of Italy, connecting the city of Turin with Venice, passing through the metropolitan area of Milan. Along this highway segment, more than $1000$ LTE cells, corresponding to about $200$ eNodeBs (eNBs), are deployed to provide connectivity to the users commuting or traveling on this path, and to citizens leaving in the proximity of the highway.
Available data exploit local monitoring information of LTE eNBs including both averaged and aggregated Key Performance Indicators (KPIs) within a time granularity of $15$ minutes. 



\subsection{Spatial consideration}
Fig.~\ref{fig:spatial_volume_distribution} (top) depicts a snapshot of normalized DL and UL traffic volumes (per single base station, where the approximate base station location is highlighted in the upper part of the plot) aggregated over a month. We highlight areas of the main cities located nearby the A4 roadway. As expected, DL volumes increase in the proximity of major cities, due to urban traffic leakage of base stations covering the highway. Additional traffic peaks can be identified in specific geographical locations, e.g., intersections between different highways, main inter-urban roads, or train lines. Generally, these locations are characterized by a higher density of base stations, which means that occurrence of any road events in these points will probably lead to major impact on the mobile network. Moreover, a large number of mobile terminals characterizes such locations over time, due to a significantly higher user mobility. Similarly, UL traffic shows analogous behaviors but with a limited volume (about $10\%$ of the overall DL traffic).


\subsection{Temporal patterns consideration}

\begin{figure}[t!]
\centering
\includegraphics[width=\columnwidth, clip, trim = 0cm 0cm 0cm 0cm]{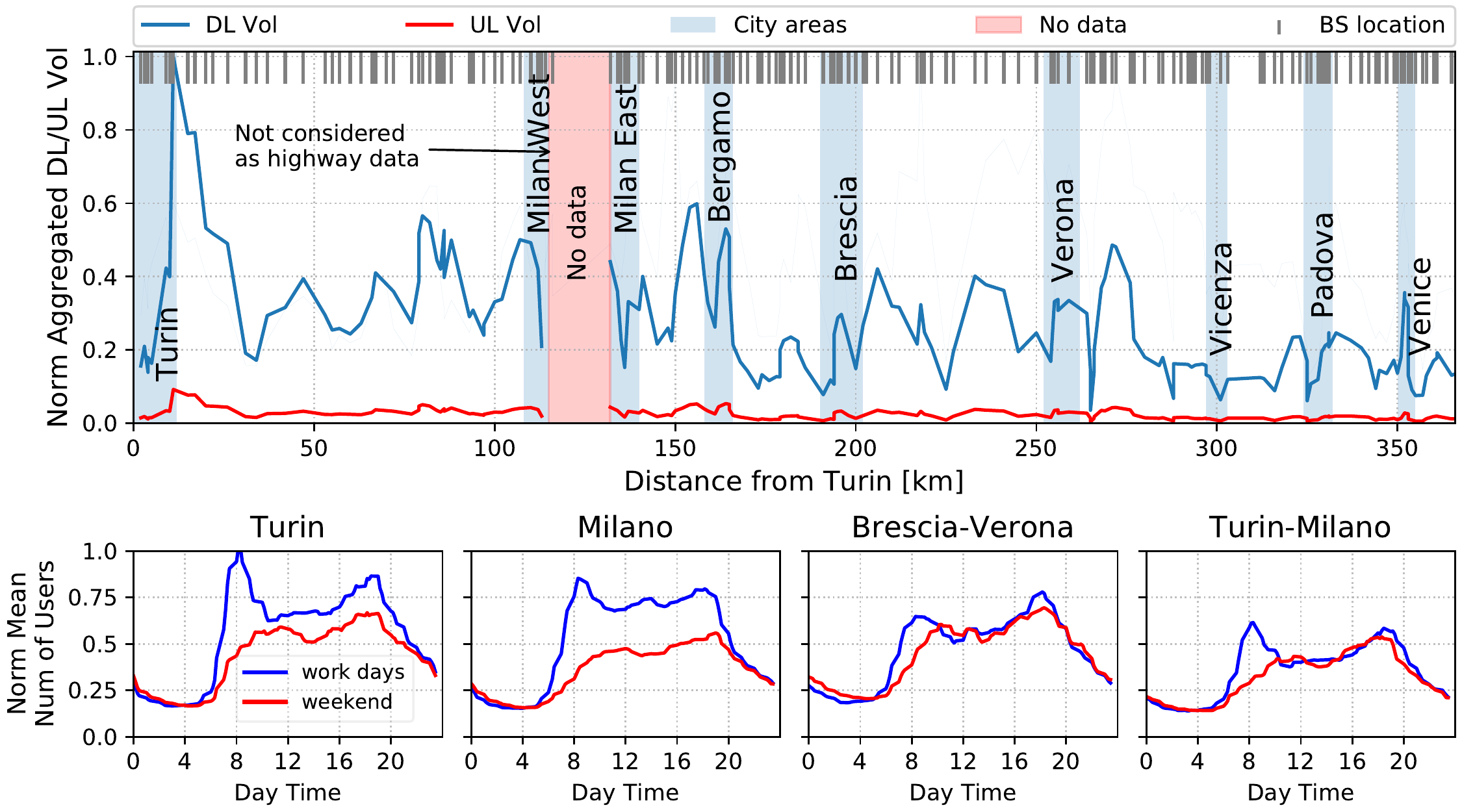}
\caption{\small Aggregated spatial distribution of DL/UL traffic volumes along the highway (top), with the focus on working day and weekend signatures for specific highway areas (bottom).}
\label{fig:spatial_volume_distribution}
\end{figure}

Several works in the literature suggest a strong relationship that correlates end-users mobility patterns with cellular network statistics~\cite{mobicom2018_MGFBC,Fiore_TMC18,okic2019WCNC} in urban environments. In a similar way, the mobility patterns identified on highways present repetitive trends following regular working routines.
In this respect, we plot at the bottom of Fig.~\ref{fig:spatial_volume_distribution} normalized number of active users during working days and weekends for different areas of the highway, in particular two areas around metropolitan cities as well as two sections of the highway between two main cities. The signatures are extracted by calculating the median values of same time periods over several weeks, separately for working days of the week and weekends, as proposed by~\cite{FurnoTMC2017}. 

The first and second subplots show that areas around big industrial cities, such as Milan and Turin, are characterized by commuting patterns with a presence of mobility peaks in the morning, noon and evening during the working days, which are absent on weekends. Conversely, mobility patterns of users during working and weekend days are comparable when considering highway segments interconnecting major cities (third and fourth subplots).



\subsection{Service-based QoS consideration}
\begin{figure}[t!]
\centering
\includegraphics[width=\columnwidth, clip, trim = 0cm 1cm 0cm 1.7cm]{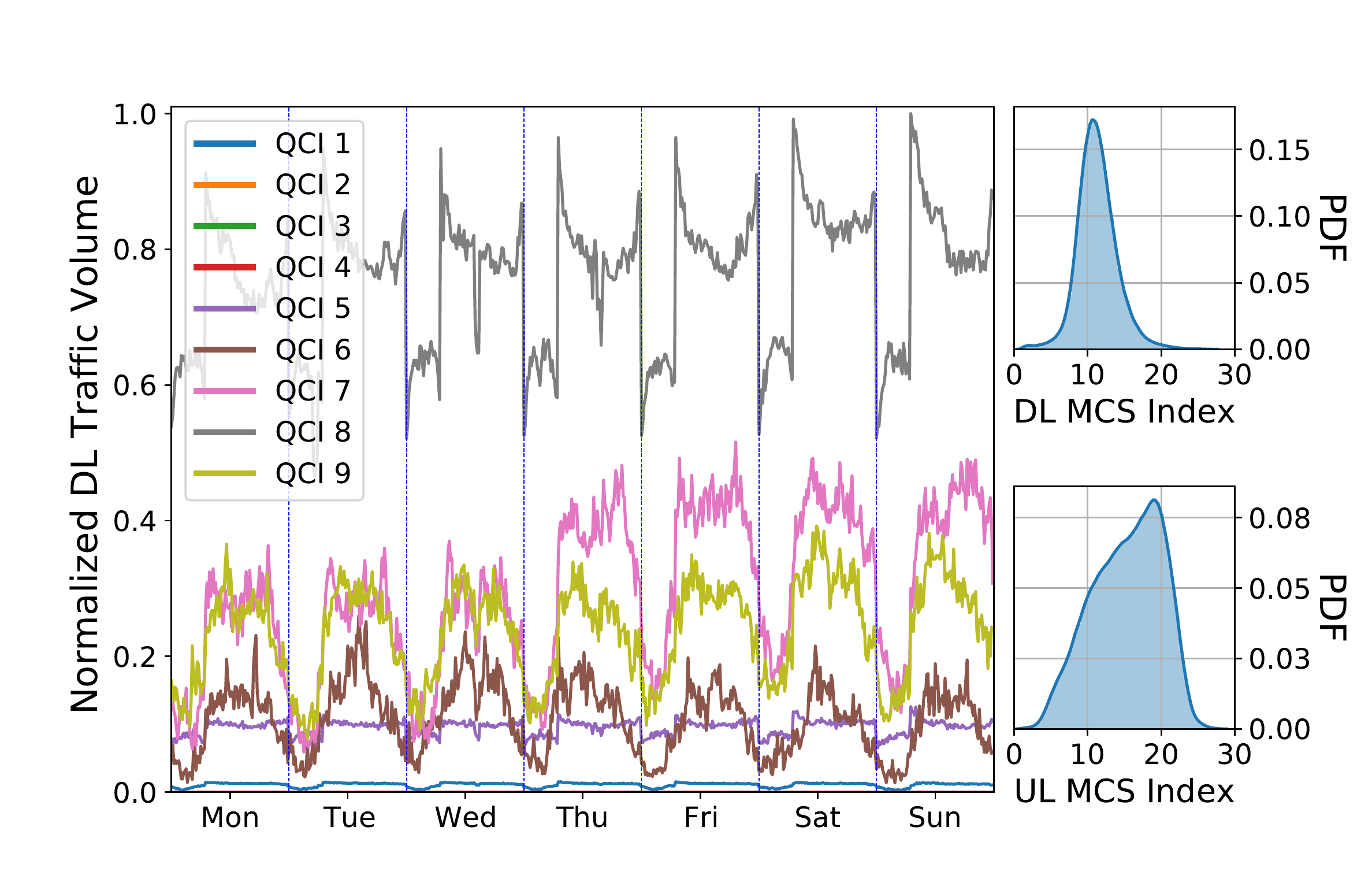}
\caption{\small Temporal distribution of different traffic types volumes.}
\label{fig:qci_mcs}
\end{figure}

Different services with heterogeneous requirements might be seamlessly managed by the mobile network infrastructure. In real deployments, this is usually achieved by labelling each traffic flow with a specific QoS Class Identifier (QCI) to ensure that each traffic bearer~\cite{TS23.303} is allocated with the appropriate set of resources to guarantee an affordable Quality of Service (QoS).
In Fig.~\ref{fig:qci_mcs} (left-hand side), we depict the temporal distribution of the DL volumes differentiated by QCI traffic types. It can be noticed a dominance of non-guaranteed bit-rate (NON-GBR) traffic types (QCIs $6$ to $9$), mostly related to video-streaming and social media activities, and almost negligible volumes for guaranteed bit-rate (GBR) traffic types (QCIs $1$ to $4$). Note that the satisfaction of service requirements also depends on the network congestion level and on the instantaneous channel quality experienced by end-users, together with the corresponding modulation and coding scheme (MCS) selected at the eNodeB scheduler. Interestingly, DL traffic perceives, on average, a lower MCS index with respect to the UL one (right-hand side of Fig.~\ref{fig:qci_mcs}). This is due to the high end-user mobility and longer data exchange sessions, which suffer from a wider communication distance along the path~\cite{Connected_Cars_IMC17}.

\subsection{Road events consideration}
\label{subsec:Traffic_event_statistics}
The considered dataset takes into account monitoring information from the mobile network infrastructure, but it lacks of information related to vehicular traffic and unexpected congestions/events that affect its dynamic.
To fill this gap and help us to accurately relate network traffic dynamics with road events, we rely on publicly available information coming from heterogeneous social media sources like Twitter notification service of \textit{Autostrade per l'Italia}\footnote{Online available at: https://twitter.com/trafficoa}, Google Maps\footnote{Online available at: https://www.google.com/maps} and Waze\footnote{Online available at: https://www.waze.com/livemap}, which facilitate an exhaustive catalog of real-time road event information and useful metadata, e.g., timestamp, exact location and a short description. 

\begin{figure}[t!]
\centering
\includegraphics[width=\columnwidth, clip, trim = 0cm 0.5cm 1cm 3cm]{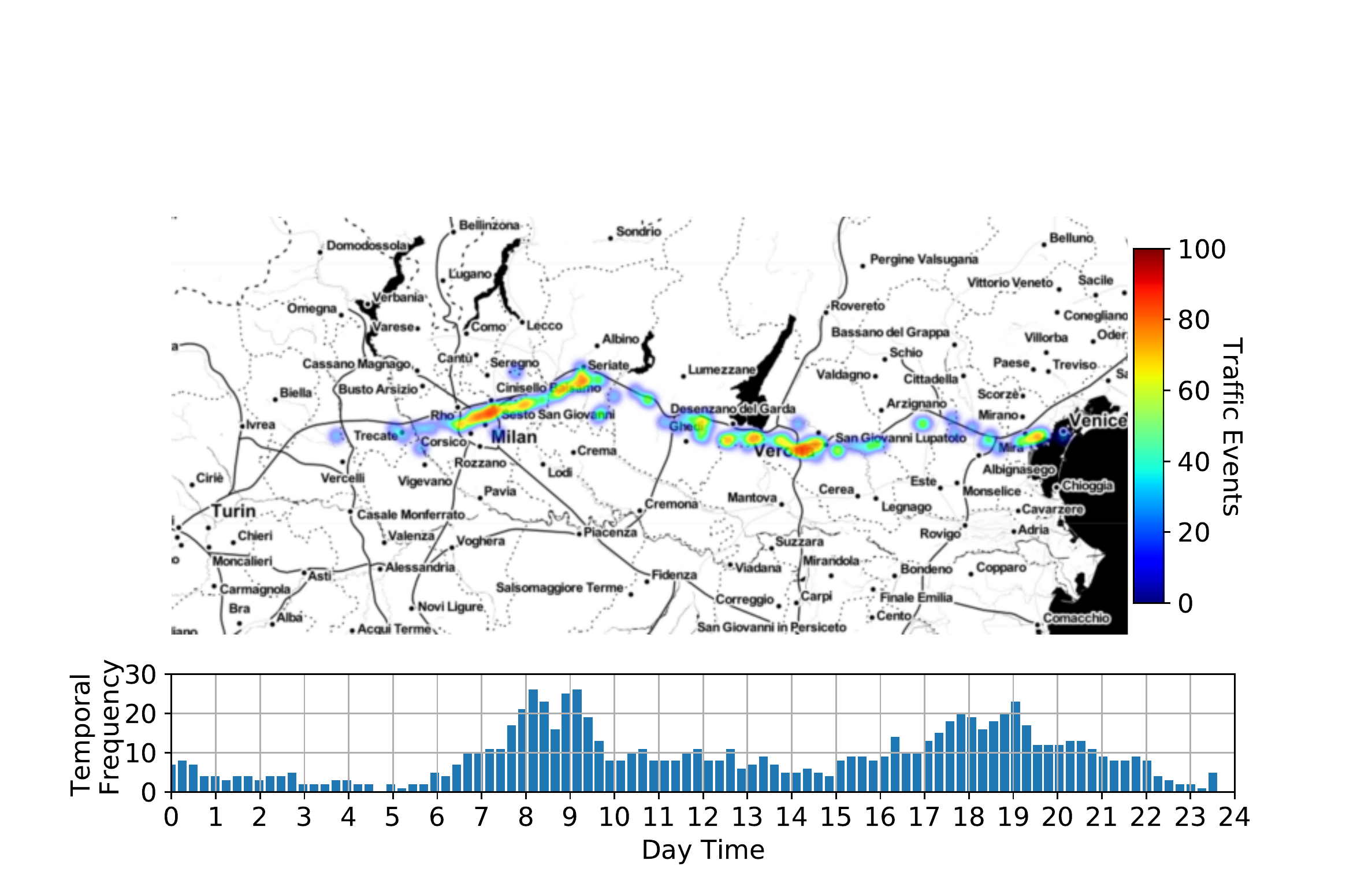}
\caption{\small Spatio-Temporal distribution of road events.}
\label{fig:events_distribution}
\end{figure}

After $6$ months of data collection, the final dataset includes about $800$ road events,\footnote{Note that our experimental campaign has been carried out during the Covid-19 pandemic, when the imposed lockdown limited the overall end-user mobility in the north of Italy, therefore, reported numbers may be biased and differ from yearly regular data.} as shown in Fig.~\ref{fig:events_distribution}. As expected, road events mainly occur during morning and evening commuting periods, and are geographically placed close to main cities, e.g., Milan and Verona.
We will match the information contained in this supplementary dataset with the network geographical deployment information to assign each road event with the closest base station and obtain the ground truth of occurred events, as later detailed in Section~\ref{sect:design}.



\subsection{Event propagation effect}
\label{subsec:propagation_effect}

Road events might have a consistent impact on the vehicular traffic conditions and, in turn, on overall network resource availability. However, it might be easily confused with common network traffic outliers and therefore be ignored. Current network deployments may come to help: the RAN deployment along the highway has quite a regular pattern and might reveal implicit information as monitoring data are simultaneously retrieved from different observation points. In particular, a road event might affect the end-users activity or drop the mobility rate gradually for adjacent base stations. We call it \emph{event propagation effect}. It usually depends on the severity of the road event, the day time, the specific location and the network deployment (e.g., base station density, etc.).

From the auxiliary information contained in our event dataset, we make a straightforward association between the location of each event and the closest base station, thereby identifying the source of the road event in the network. Then, analyzing the propagation of anomalous traffic patterns onto nearby cells, we can infer which direction of the highway, or vehicular traffic flow, has been affected. 

\begin{figure}[b!]
\centering
\vspace{-5mm}
\includegraphics[width=\columnwidth,clip, trim = 0cm 0cm 0.7cm 1cm]{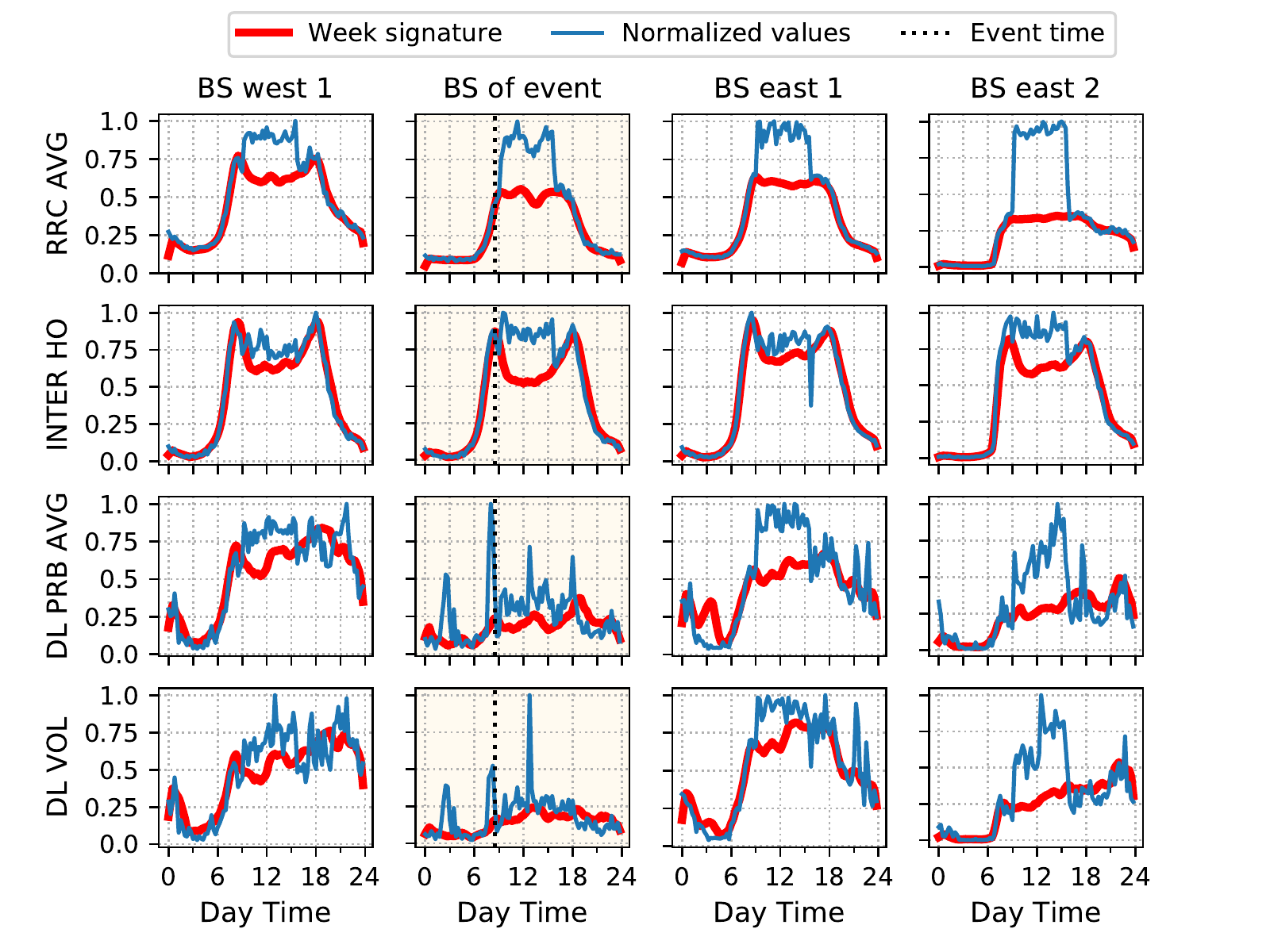}
\caption{\small Event propagation on nearby base stations in both directions for selected metrics.}
\label{fig:event_propagation}
\end{figure}

An example of this is shown in Fig.~\ref{fig:event_propagation}. The scenario accounts for a major accident occurred nearby Verona at $9$:$00$ AM. The red line in each plot represents the regular behaviour of the corresponding metric, while the blue line depicts the daily data. After the accident, we can notice a clear deviation from regular patterns for multiple metrics lasting few hours. The higher number of users in RRC connected state indicates that multiple users are actively using mobile network resources, suggesting the presence of traffic congestions. Similarly, this affects DL volumes and average Physical Resource Block (PRB) utilization. Moreover, a higher number of affected base stations on the east side suggests that the accident occurred in east-west direction (Venice-Turin).
Given the high variability and intrinsic characteristic of each monitoring metric, an accurate anomalous pattern characterization would involve time, space and metric-specific considerations. Therefore, road events can hardly be detected by simple anomaly detection algorithms, e.g., those based on outlier detection that may run on a single base station. This motivates us to investigate advanced deep learning solutions capable of dealing with
such multi-dimensional information matrix while keeping a global view of the system.

\begin{figure*}[t!]
\centering
\includegraphics[width=\textwidth, clip, trim = 0cm 0cm 0cm 0cm]{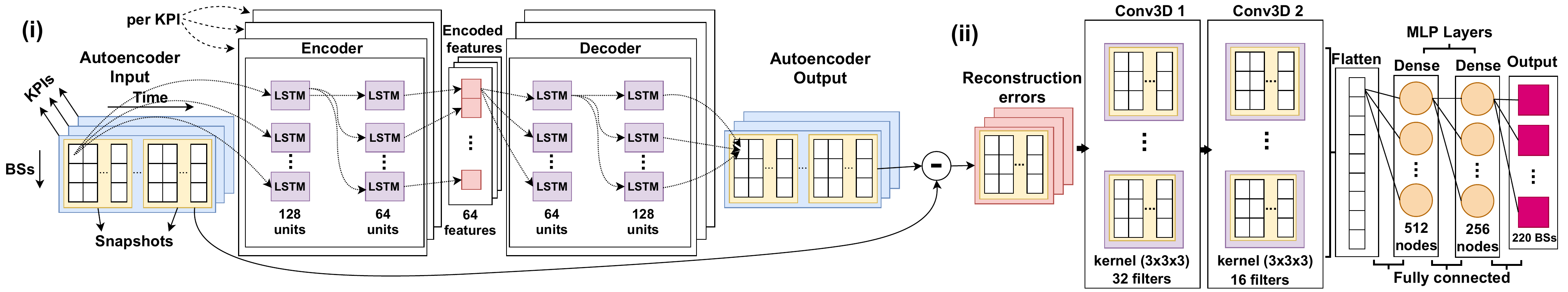}
\caption{\small \name{} model architecture}
\label{fig:model_architecture}
\end{figure*}

\section{ \name{} Design and Model Formulation}
\label{sect:design}
Hereafter, we introduce the design of \name{}, a deep learning-based framework to detect road events. An overview of the model architecture is depicted in Fig.~\ref{fig:model_architecture}. The overall solution requires two different stages to deal with road event predictions: $i$) an autoencoder-based approach consisting of Long Short-Term Memory Recurrent Neural Networks (LSTM) layers to detect anomalies within highly dynamic and heterogeneous time patterns and $ii$) a Deep-Learning approach to approximate complex functions, e.g., those that relate road events occurrences with temporal and spatial distributions of anomalies detected in adjacent base stations. The overall $6$ months dataset has been splitted according to a $60/20/20$ ratio for the purposes of training, validation and testing procedures, respectively. 


\subsection{Input of \name{}}
Let us consider a time slotted system whereby time is divided into time intervals $t=\{1, \dots, \mathcal{T}\}$, and define $\mathcal{N}$ as the set of base stations deployed along the highway. As detailed in Section~\ref{sect:data_analysis}, each base station $n \in \mathcal{N}$ gathers information about a multitude of heterogeneous network statistics, defined as $\mathcal{M}$, which include, among the others, PRB utilization, physical channel quality information, traffic volumes, X2/S1 hand-over statistics, etc. In our experiments we consider $25$ different monitoring metrics.

Let $x^n_m(t)$ be the time series describing the $m$-th monitoring metric observed under the base station $n$ at the time $t$. As suggested in~\cite{Wang_TMC2019}, an accurate organization of the input traces helps enhancing deep learning models performances. In light of this, for all metrics $m \in \mathcal{M}$ and base stations $n \in \mathcal{N}$, we collect the $x^n_m(t)$ traces and order them by preserving their spatial location (from Turin to Venice) in a matrix $X_m(t)$ with dimensions $N \times L$, representing a metric-specific \emph{snapshot} of the network, where $N$ is the total number of base stations, and $L$ is the number of monitoring samples within the observation period (or \textit{lookback} time window), fixed to $4$ hours throughout our experiments. The input matrices $X_m(t)$ are normalized with respect to maximum values of each $m$ in the training dataset.

\subsection{LSTM-based anomaly detection}



The possibility to detect changes in traffic conditions is subject to the capabilities of our model to correctly identify irregular statistics from the monitoring samples collected along the highway. Given the multi-dimensionality of our dataset, we build an autoencoder $A_m$ for every collected metric thereby allowing for better scalability.
Autoencoders imply the setup of an encoding-decoding architecture. The encoding part, commonly implemented as feed-forward neural networks, provides a compressed representation of the input data to subsequent layers~\cite{Auxiliary_Information}.
The decoding phase follows the same steps over a symmetric architecture. Analytically, the two phases applied to the input metric $X_m$ can be described as follows:
\begin{align}
    h_m(X_m(t)) = \eta(W^\eta_m X_m(t) + b_m^\eta), \\
    \hat{X}_m(t) = \delta(W^\delta_m h_m(X_m(t)) + b_m^\delta),
\end{align}
where $\eta$ and $\delta$ are the encoding and decoding functions with their corresponding weights $W^\eta_m$ and $W^\delta_m$, $b_m^\eta$ and $b_m^\delta$ are bias vectors, and $h_m(X_m(t))$ and $\hat{X}_m(t)$ are the compressed input and the reconstructed output sequences, respectively.

The performance of autoencoders, defined as their capability to reconstruct the input sequence, depends on variability and complexity of data provided at the input. Differently, in our work we exploit autoencoders for anomaly detection~\cite{Detecting_Anomalies} instead of simple dimensionality reduction.

The data analysis performed in Section~\ref{sect:data_analysis} suggests the adoption of deep learning techniques able to deal with the spatio-temporal characteristics of mobile traces. For this reason, we implement our encoder-decoder architecture by means of Long Short-Term Memory Recurrent Neural Networks (LSTM-RNNs). Each autoencoder $A_m$ accounts for $4$ LSTM layers, two for each phase. LSTM is a type of RNN architecture that has proven its value when dealing with repetitive patterns and unstructured time series, while solving a problem of vanishing gradient for long-term dependencies present in other RNNs~\cite{LSTM}. LSTMs can be represented as a chain of $G$ modules, or cells, each one applying a set of operations to the input data. In our case, the two LSTM layers are composed by $128$ and $64$ cells, respectively~\cite{LSTM_AE_trinh2019detecting}.

The output of each cell $g \in G$, also known as cell state, is transferred to the subsequent cell in a recursive manner. The possibility of handling long-term trends in LSTM is provided by structures, dubbed as gates, which carefully remove or add information to the cell state. Each cell has three gates, namely, input $I_g$, output $O_g$ and forget gate $F_g$, which controls the amount of information that should be added (or dropped) to the cell state before transferring it to the next unit~\cite{LSTM_what_to_do_next_2017}.
This effect is achieved combining the influence of different non-linear activation functions, i.e., $\sigma$ (sigmoid) and $\tanh$ (hyperbolic tangent function), at each gate. The impact of these functions on the input data needs to be learned during the training phase aiming to minimize a loss function~\cite{LSTM_what_to_do_next_2017}, which in our case is the mean squared error (MSE).

To accomplish the anomaly detection task, we make use of labeled data and train the autoencoders offline exclusively on \emph{eventless} snapshots taken from historical data as to capture the behavior of the system without anomalies. Once trained, we feed the model with online monitoring traces. Given that anomalous patterns have not been part of the training phase, it is expected that the model will exhibit performance degradation during the reconstruction phase. Simple classifiers would mark the input snapshot as anomalous if the reconstruction error exceeds a given threshold. Instead, we derive the squared error matrix $e_m(t)= \|{X_m(t)-\hat{X}_m(t)}\|^2$ from each autoencoder $A_m$ and combine the individual error matrices into a 3D tensor $e(t) = \{e_1(t), e_m(t), \dots, e_M(t) \}$  which is passed to the second stage of our model.

\subsection{Road Event Localization}
\label{sect:event_local}
The function linking anomalies detected by the first stage of our model with the occurrence of road events along the highway is unknown and hard to be characterized due to the multitude of system-related and external variables affecting the entire detection process. The simple detection of anomalies in monitoring statistics does not imply the occurrence of an emergency, as uncorrelated events---like hardware failures and/or unexpected traffic peaks---may trigger alerts leading to erroneous estimations. For this reason in the following, we exploit the information described in Section~\ref{subsec:Traffic_event_statistics} as ground-truth to train a Convolutional Neural Network (CNN)-based model that captures the spatio-temporal correlation of different anomalies and maps them into geographical information~\cite{3D_conv_ji20123d}. Our design choice is further motivated by the fact that in case of road events, as shown in Fig.~\ref{fig:event_propagation}, affected base stations present significant levels of correlation between each other, which further improves the learning task.

More in details, the second stage of \name{} consists of two stacked 3D-CNN layers and a final Multi-Layer Perceptron (MLP) fully connected layer.
In order to exploit local correlation from nearby base stations, each neuron of the 3D-CNN layers has a limited \textit{receptive field}, or kernel, whose size determines its observation area. For a given input tensor, convolutions with 3D kernels are iteratively applied to provide the subsequent layers with a compressed representation of the input information.

Through extensive hyper-parameters optimization, we used two 3D-CNN layers with filter sizes of $32$ and $16$ neurons, respectively, and corresponding convolutional kernel sizes of $(3,3,3)$ and $(3,3,3)$~\cite{3DCNN_yang2019imgsensingnet}.
Each neuron of the CNN runs a filter $H(\sum_t e(t)*k(t) + b)$, where $e(t)$ is the input tensor at time $t$, $k(t)$ is the kernel filter, $*$ is the convolution operator, $b$ is a bias, and $H(\cdot)$ is a non-linear activation function, in our case Rectified Linear Unit (ReLU)~\cite{ML_zappone2019wireless}.

In order to map the encoded representation of the anomalies into geographical information about the emergency, we make use of a final Multi-Layer Perceptron layer. MLP is a class of neural networks with fully-connected neurons among layers which has the capabilities to approximate,
through supervised learning, the function that relates the input with the output. In our case, we are interested in the function that links the encoded spatio-temporal anomalies of multiple metrics and different base stations with the exact event location. The MLP classifier consists of three layers with $512$, $256$ and $220$ neurons, respectively, where the output layer matches the number of base stations~\cite{CNN_MLP_sankhe2019oracle}. To regularize the output and reduce over-fitting, we introduce a dropout rate of $0.2$. Overall, the second stage of the model is trained using Adam optimizer with learning rate $10^{-4}$, and adopting cross-entropy as loss function over $150$ training epochs.
Due to high system variability caused by user social behaviors and external causes affecting it (e.g., weather conditions), it clearly rises the need to design a model capable of adapting to new (anomalous) patterns. Therefore, we ensure that the model is retrained as soon as new observations are made available.

\subsection{Road Event Classification}
\label{subsec:event_classification}

Upon detection, it is important to classify the magnitude of each event to promptly alert first aid responders and, if necessary, identify the set of networking requirements to be allocated for an emergency slice setup. We rely on the impact of the \emph{propagation effect} over the set of base stations near to the road event to provide an empirical classification. 
Let us introduce $\Tilde{\mathcal{N}} \subset \mathcal{N}$ as the set of base stations affected by the road event.
Due to the irregular density that characterizes the Radio Access Network (RAN) deployment along the highway, we argue that a simple road event classification based on the cardinality of $\Tilde{\mathcal{N}}$ would not be accurate, as road events occurring in the proximity of main cities would involve a wider number of base stations than those occurring in rural areas. Therefore, we define our classification metric $\mu$ as $\frac{\Tilde{N}}{\psi} $, where $\Tilde{N}$ is the cardinality of $\Tilde{\mathcal{N}}$, and $\psi$ is the number of base stations deployed in the area within a radius of $10$~km surrounding the road event. We empirically select $10$~km as this value represents the longest vehicular queue registered in our event dataset.
Clearly, $\psi$ should be re-dimensioned to generalize our findings within other network topologies.


Finally, we differentiate among three different categories of events: \textit{Light}, \textit{Moderate}, and \textit{Severe}. The event duration of each category reflects the most common scenarios contained in our real dataset, while networking throughput requirements are meant to support the provisioning of Augmented and Virtual Reality (AR/VR) streaming in mission critical scenarios~\cite{Mission_Critical_slicing}. The details about each road event class and corresponding ENS requirements are summarized in Table~\ref{tab:ENS_requirements}.

\begin{table}[b]
\footnotesize
\centering
\caption{\small Event characterization and ENS Requirements.}
\label{tab:ENS_requirements}
\resizebox{\columnwidth}{!}{\begin{tabular}{|c|c|c|c|c|}

\hline
\rowcolor[HTML]{EFEFEF}
\textbf{Event category} & \textbf{$\Theta$ - Time Duration} & \textbf{NS Requirements}  & \textbf{$\mu$ Value} \\ \hline 
Light           & $20$min      & $10$Mbps          & $(0,1]$\\ 
Moderate        & $40$min      & $15$Mbps          & $(1,2]$\\ 
Severe          & $60$min      & $25$Mbps          & $\geq2$   \\ \hline

\end{tabular}}
\vspace{-4mm}
\end{table}

\subsection{Performance Comparison and Practical Considerations}
\label{subsec:practical_considerations}

Hereafter, we compare the performance of \name{} against state-of-the-art models trained to perform similar anomaly detection and classification tasks, highlighting each model drawback (compared to our solution) through practical considerations. 
Benchmarks include a simple threshold-based autoencoder classifier (AE)~\cite{LSTM_AE_trinh2019detecting}, and a more advanced 3D-CNN-based classifier~\cite{3DCNN_yang2019imgsensingnet, CNN_MLP_sankhe2019oracle}.
\begin{figure}[t!]
\centering
\includegraphics[width=\columnwidth, clip, trim = 0cm 0cm 0cm 0cm]{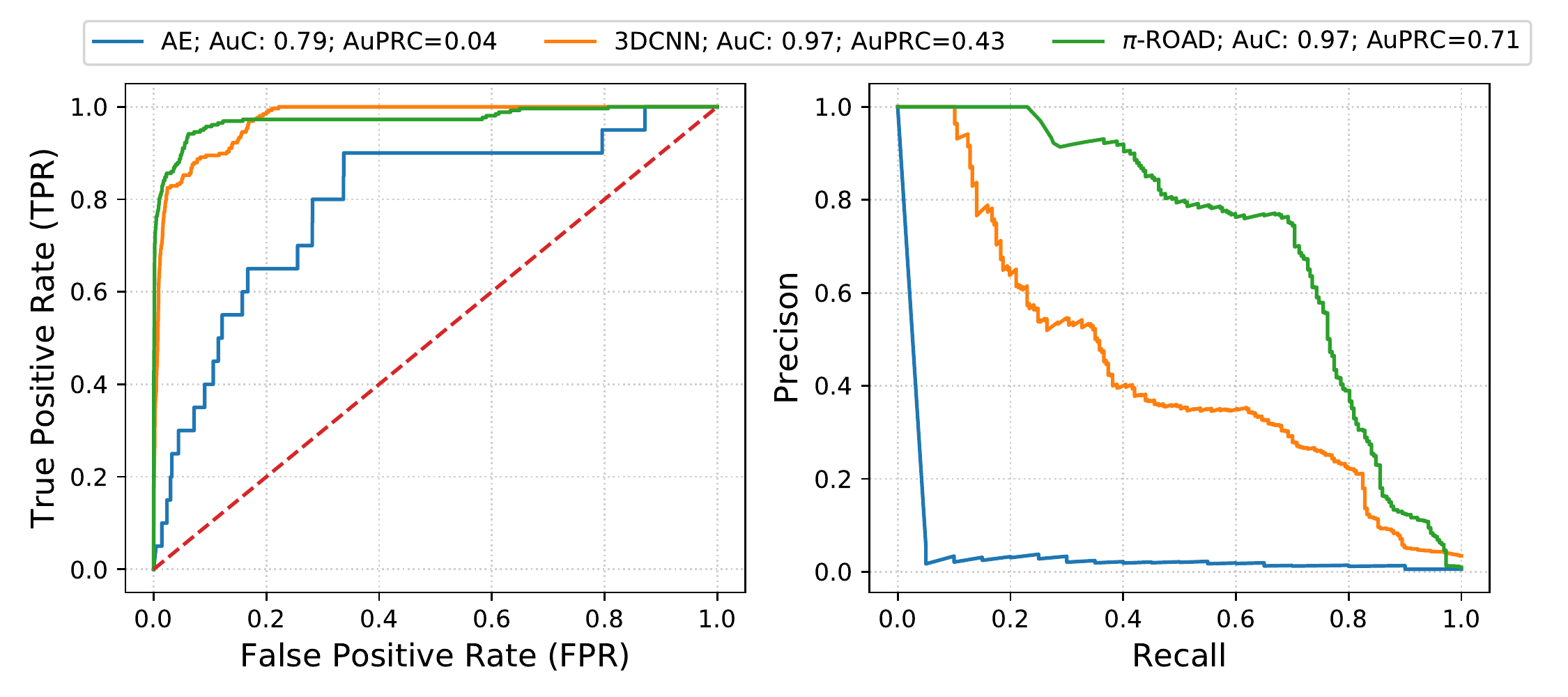}
\caption{\small ROC and PR diagrams with indicated AuC and AuPRC scores for different anomaly detection and classification approaches.}
\label{fig:roc_pr_diagram}
\end{figure}
Fig.~\ref{fig:roc_pr_diagram} resumes the overall performances using Receiver Operating Characteristics (ROC) and Precision-Recall (PR) metrics. On the left-hand side, the ROC metric provides a compact representation of the capabilities of the model to deal with the trade-off between True Positive Rate (TPR) and False Positive Rate (FPR).
On the right-hand side, the PR curve states the performance of a classifier in terms of Precision and Recall, where Precision is a measure of result relevancy, and Recall is a measure of how many truly relevant results are returned by the classifier. To ease the comparison, we also quantify the Area under the ROC Curve (AuC) and Area under the PR Curve (AuPRC).
The unsupervised approach adopted in AE requires the definition of a-priori thresholds to mark anomalous patterns from the reconstruction errors. Clearly, the definition of these settings depends on the intrinsic variability of the traces in input, and requires considerations over the statistics of each feature. This approach hardly scales when considering multiple heterogeneous metrics, as in our case. Despite accurate tuning of the parameters, it clearly appears how this simple approach fails to reveal the majority of events, achieving the lowest precision score. This further emphasizes how detection schemes based on simple threshold/outlier detection are not enough to address the scenario considered in our work.
Conversely, the 3D-CNN classifier adopts a supervised approach, which prevents from specifying user-defined thresholds for decision-making. The resulting ROC and PR curves show better performances when compared against the baseline AE approach. This result can be explained through the ability of 3D-CNN network to capture spatio-temporal correlations between different network measurements~\cite{ARENA}. 
However, when Recall value increases over a certain level (i.e., moving from left to right on the PR diagram), the Precision score drops drastically, suggesting poor performances when differentiating among different types of anomalies. In other words, the model detects only some types of anomalies, e.g., those deriving from major events with very strong impact on monitoring traces, and fails to generalize over probably smaller ones.
Conversely, \name{} outperforms the two stand-alone approaches in both ROC and PR metrics. The advantage deriving from the two-stage approach adopted by \name{} is two-fold $i$) the initial anomaly detection task performed by first stage of the model, together with an accurate input organization, facilitates the learning task of the 3D-CNN layer and $ii$) the capabilities of the 3D-CNNs to deal with bare reconstruction errors removes the need to provision user-defined thresholds which may bias the final results.

\section{ENS orchestration}
\label{sect:framework}

The outcome of \name{} can be used to tackle a variety of public safety issues along the highway. For example, in case of road accidents, it is important to enable dynamic RAN resource allocation in order to provide first responder teams with enough communication capabilities, regardless of active user sessions or services. To this aim, in what follows we introduce our formulation to address the Emergency Network Slice (ENS) orchestration problem.

\textbf{Radio Access Network.} Let us consider a RAN deployment with network slicing support covering the area surrounding the highway and comprising a set of base stations $\mathcal{N}$. Each base station $n\in\mathcal{N}$ is characterized by a capacity $C_n$, defined as the sum of its available physical resource blocks (PRBs). As result of the road event localization process executed by \name{} and described in Section~\ref{sect:event_local}, we identify the subset of base stations $\widetilde{\mathcal{N}} \subset \mathcal{N}$ that will host the emergency slice.

\textbf{Active slice services.} Let us assume a set of network slices $\mathcal{S}$ supporting V2X services, being already installed and active on the considered RAN deployment, whereas let us mark $s=0$ the Emergency Network Slice (ENS) to be temporarily installed. 
We assume each V2X slice $s\in\mathcal{S}$ described by means of a predefined network slice template that suggests slice requirements in terms of throughput $\Delta_s$, and latency $\Lambda_s$. Typical values for V2X services are shown in Table~\ref{tab:v2x_slice_requirements}. 

\begin{table}[b]
\footnotesize
\centering
\caption{\small V2X Slice requirements (c.f.~\cite{TR22.886, TR22.891,Network_Slicing_for_V2X})}
\label{tab:v2x_slice_requirements}
\resizebox{\columnwidth}{!}{\begin{tabular}{|c|c|c|c|}

\hline
\rowcolor[HTML]{EFEFEF}
\textbf{V2X category}               & \textbf{Latency} & \textbf{Data rate}                   & \textbf{Reliability} \\ \hline 
Autonomous driving                  & $10$ms              & $10$Mbps                               & 99.999\%             \\ 
Tele-operated driving               & $20$ms             & $25$Mbps Sensor Data Streaming & 99.999\%             \\ 
Vehicular Internet/Infotainment & $>100$ms            & $15$Mbps Video Streaming           & Not Specified        \\ 
Road Safety      & $100$ms            & $1$Mbps                              & 99\%     \\ \hline
\end{tabular}}
\end{table}

\subsection{Problem design} 
In real scenarios, when setting up an emergency slice, advanced orchestration operations are required to still guarantee service level agreements (SLAs) of active slices. Assuming that an admission control process has been executed to accept and install V2X slices on the network with the aim of maximizing the resource efficiency (while still honouring expected service requirements), it might appear challenging to add on top of active slices an additional high-priority service, such as ENS.
However, an efficient admission control will accommodate slices with heterogeneous requirements to compensate unexpected slice behaviors or traffic peaks~\cite{DeepCogInfocom19}. 
Analytically, we assume that predefined SLAs include for each slice $s\in\mathcal{S}$ and base station $n\in\mathcal{N}$ a minimum number of reserved PRBs, namely $Q^{(t)}_{n,s}$\,\footnote{When SLAs disclose information on expected throughput or user rate, the amount of required PRBs could be dynamically obtained by monitoring the traffic demand of the different slices. For further details, we refer the reader to~\cite{Slicing2018CoNEXT,AztecInfocom20}.}. Without loss of generality, we assume that the overall resource availability is assigned to the set of running slices. 
In addition, we translate the slice latency requirements $\Lambda_s$ into a tolerance value $\lambda_s$ that defines the maximum number of time intervals packets shall wait into the queue before being served, or dropped due to lack of resource availability within the latency requirements.
Note that if all running V2X slices have high priority, i.e., very low delay tolerance, it might be infeasible to install the ENS without impairing other active services.
ENS parameters depend on the severity of the emergency road event which can be obtained as output of a \name{} execution. We resume the corresponding settings in Table~\ref{tab:ENS_requirements}, in light of the discussions of Section~\ref{subsec:event_classification}.
We define $\Theta$ as the envisioned emergency time duration, and assume, for each ENS to be deployed, a fixed amount of PRBs allocation $Q_{n,0}^{(t)}, \forall t\in \{0,\cdots,\Theta\}$ and the lowest delay tolerance $\lambda_0=1$.
If not properly scheduled, the additional ENS may lead in the worst case to resource deficit and service disruption in one or multiple base stations. Therefore, we aim at minimizing the instantaneous resource deficit $\pi_n^{(t)}$, while still guaranteeing defined SLAs for other active running slices. Our problem can be formulated as follows:

%
\begin{problem}[$\pi$-Orchestrator]\label{eq:problem_slice_orch}
\begin{align}
   & \mathrm{minimize} && \displaystyle \sum_{n\in\Tilde{\mathcal{N}}} \sum_{t\in\mathcal{T}} \pi_n^{(t)} & \nonumber\\
   & \mathrm{subject~to} && \sum_s z_{n,s}^{(t)} \leq C_n + \pi_n^{(t)}, \quad\quad\quad \forall n\in\Tilde{\mathcal{N}},t\in\mathcal{T}; \nonumber\\
   & && \sum_{t=0}^{\Theta} Q_{n,s}^{(t)} - z_{n,s}^{(t)} \leq \sum_{t=\Theta+1}^{\Theta+\lambda_s} Q_{n,s}^{(t)}, \, \forall s\in\mathcal{S},n\in\Tilde{\mathcal{N}}; \nonumber\\
   & && z_{n,0}^{(t)} = Q_{n,0}^{(t)};\quad \forall n\in\Tilde{\mathcal{N}},t\in\mathcal{T}; \nonumber\\
   & && z_{n,s}^{(t)} \in \mathbb{N}^+, \qquad \forall n\in\Tilde{\mathcal{N}},s\in\mathcal{S},t\in\mathcal{T}; \nonumber\\
   & && \pi_{n}^{(t)} \in \mathbb{N}^+, \qquad\forall n\in\Tilde{\mathcal{N}},t\in\mathcal{T}. \nonumber
\end{align}
\end{problem}
\noindent where (integer) decision variables are $\pi_n^{(t)}$ and $z_{n,s}^{(t)}$ indicating the number of PRBs to be assigned to slice $s$ on base station $n$ at time interval $t$. Note that $z_{n,0}^{(t)} = Q_{n,0}^{(t)}$ is due to the highest priority assigned to the ENS ($s=0$). The first set of constraints introduces flexibility into the problem by adding a non-zero fictitious value (i.e., the resource deficit $\pi_{n}$) to avoid infeasible solutions. The second set of constraints specify that the slice resource reservation must be performed fulfilling the slice delay tolerance, i.e., waiting traffic still in the queue must be scheduled within $\lambda_s$ time slots. We run Problem~\ref{eq:problem_slice_orch} for a set of time slots $\mathcal{T}$ that includes the time window $\Theta$ required by the ENS.
Problem~\ref{eq:problem_slice_orch} is an Integer Linear Programming (ILP), which can be efficiently approximated by means of relaxation techniques (e.g.,~\cite{Lagrangian_2008}), and commercial solvers to provide near-optimal results\,\footnote{The implementation of the problem has been carried out using the framework of {\ttfamily IBM ILOG CPLEX} and its {\ttfamily Python API}.}.

\begin{proposition}
Problem~\ref{eq:problem_slice_orch} is NP-Hard and difficult to approximate within $N^{1-\epsilon}$, for $\epsilon>0.$
\end{proposition}

\begin{IEEEproof}
The NP-Hardness proof goes by reduction. Problem~\ref{eq:problem_slice_orch} clearly belongs to the NP problems. Let us consider an instance of a bin packing problem (BPP) with arbitrary constants $a$ and $B$, and binary decision variables $x_{ij}$ and $y_j$, where $\mathcal{I}$ and $\mathcal{J}$ are the sets of items and bins, respectively~\cite{Martello_Knapsack}.

\begin{problem}[BPP]\label{eq:problem_BPP}
\begin{align}
   & \mathrm{minimize} && \displaystyle \sum_j y_j & \nonumber\\
   & \mathrm{subject~to} && \sum_i ax_{ij} \leq B y_j, \quad\quad\quad \forall j\in\mathcal{J}; \nonumber\\
   & && \sum_j x_{ij} \leq 1, \quad\quad\quad \forall i\in\mathcal{I}. \nonumber
\end{align}
\end{problem}

If we consider $N=1$ base station, and $z_{n,s},\pi_n$ as binary decision variables, Problem~\ref{eq:problem_slice_orch} reduces to a BPP that shows that for all $\epsilon > 0$, packing items within the minimum number of bins within $N^{1-\epsilon}$ is NP-hard~\cite{Martello_Knapsack}. Since such a problem is trivial compared to our original Problem~\ref{eq:problem_slice_orch}---that includes a number of base stations $N>1$ and integer decisions variables---this result is rather strong. 
\end{IEEEproof}

\subsection{V2X traffic scheduling}
The instantaneous resource deficit $\sum_{n\in\Tilde{\mathcal{N}}} \pi_n^{(t)}$ depends on the severity of the detected road event and, in turn, on the specific ENS resource reservation and time duration settings. Therefore, Problem~\ref{eq:problem_slice_orch} considers the worst case scenario, i.e., when each active 
slice fully utilizes reserved resources. However, as shown in Section~\ref{sect:data_analysis}, network congestions rarely occur outside the commuting time periods and some of reserved network resources may be underutilized. So, the impact of the ENS on the system can be further mitigated relaxing the fixed PRB allocation scheme envisioned to assure slice isolation at RAN scheduling level~\cite{foukas_orion, LACO}.
Specifically, if a network slice is underutilized, other slices with pending traffic can use some of its network resources (PRBs)\,\footnote{While this concept may fail to comply with the network slice isolation principle, we assume that the isolation is still guaranteed at higher (abstracted) scheduling layers~\cite{foukas_orion}.}.
This allows us to devise a practical algorithm that reduces the overall slice resource deficit. In particular, the algorithm sorts slice traffic requests based on the slice priority (or traffic delay tolerance). Slices with higher priority will use assigned PRBs as per the solution of Problem~\ref{eq:problem_slice_orch} ($z_{n,s}^{(t)}$) within time interval $t$ to serve incoming traffic requests. Once all requests have been served, remaining slice PRBs ($z_{n,s}^{(t)}>0$) are used to serve traffic requests of the next slice in the priority-ranked list. When all PRBs are used ($\sum_s z_{n,s}^{(t)}=0$), the algorithm proceeds to the next time interval $t$ keeping unserved traffic requests in the slice queues. If slice traffic requests are not served within the slice delay tolerance $\lambda_s$, these are dropped.
The resulting resource deficit is further minimized as the RAN slice scheduler efficiently assign resources to slices with pending traffic. Results are shown and discussed in Section~\ref{sect:perf_res_orch}.

\section{ \name{} Performance Evaluation}
\label{sect:perf_eval}

In the following, we investigate the performance of \name{} while detecting road events, and use this information to assess our solution's capabilities in orchestrating radio resources in case of road accidents and realistic traffic conditions involving different types of V2X network slices.

\subsection{Localization of the road event}
\begin{figure}[t!]
\centering
\includegraphics[width=\columnwidth, clip, trim = 0cm 0cm 0cm 0cm]{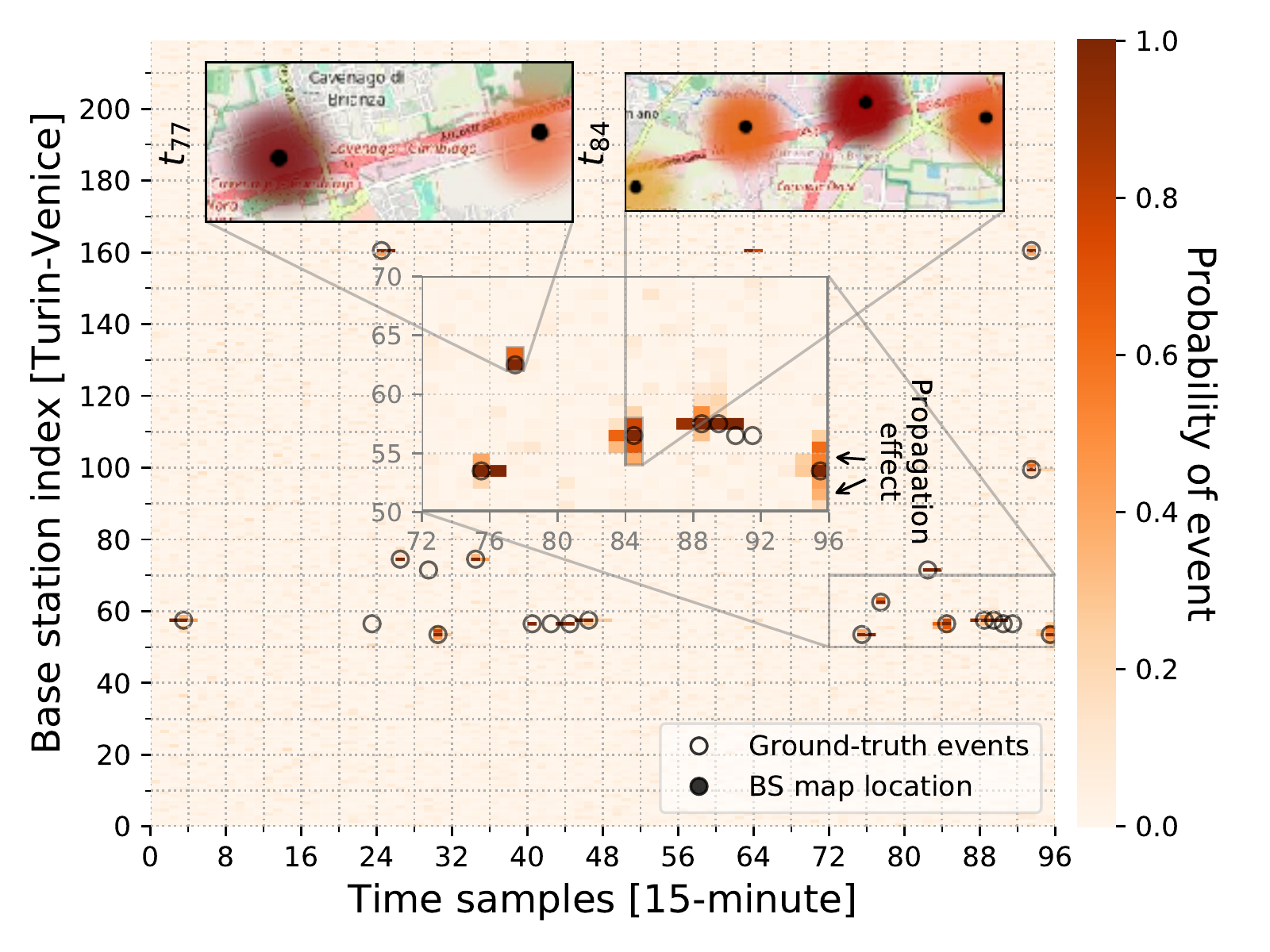}
\caption{\small Example of road event detection and localization using \name{} model along the highway within one day, compared against the event ground-truth information.}
\vspace{-3mm}
\label{fig:heatmap_pred_vs_real}
\end{figure}

In order to evaluate event localization performances, we provide as input the test set of monitoring samples taken from our dataset. In Fig.~\ref{fig:heatmap_pred_vs_real}, we summarize the results, focusing on the set of base stations and a representative time period. 
Each point on the map represents a pair of base station and time interval, and its color indicates the detected probability of event occurrence given the monitoring information. Circular markers on the same map highlight the ground-truth location of road accidents.
From the picture it can be noticed how \name{} detects most of the events, not only revealing their temporal duration (x-axis), but also 
suggesting how much the propagation effect influences adjacent base stations (y-axis).
Interestingly, the heatmap also reveals some cases in which \name{} detects the occurrence of events before online notification services and social media platforms. For visualization purposes, at the top of the figure we highlight two geographical areas of the highway covering impacted base stations in two different event occurrences happening at time $t_{77}$ and $t_{84}$, respectively. In the first case, \name{} classifies the road event as \textit{Light} since only two base stations reported anomalous statistics. In the second case, given the wider number of affected eNodeBs, the event is classified as \textit{Moderate}.
Finally, few events are not detected. Either this can be due to very limited impact of such events on the monitoring traces or, they may be approximated with patterns that are unknown to the model. In the latter case, we expect that similar events would be detected in the future as the model follows the learn-as-you-go approach to continuously train with the latest data.

\subsection{Resource Orchestration}
\label{sect:perf_res_orch}
Hereafter, we test our ENS orchestration solution exploiting \name{}'s outputs and the realistic monitoring information contained in our dataset.
We consider a $4$-slices network scenario, including two enhanced Mobile Broadband (eMBBs) slices dedicated to streaming and infotainment services (with high delay tolerance), and two Ultra-reliable low-latency communication (URLLC) slices dedicated to Autonomous and Tele-operated driving, with corresponding networking requirements as detailed in Table~\ref{tab:v2x_slice_requirements}. We assume that DL slice traffic volumes are generated under the reference set of base stations in the considered event period, according to the specific service QCI and corresponding traces as depicted in Fig.~\ref{fig:qci_mcs}. In particular, eMBB slices are mapped to QCIs traces $7$ and $8$, while URLLCs to QCIs $1$ and $6$, respectively. 


\begin{figure*}[t!]
    \centering
  	\begin{subfigure}[t]{.5\textwidth}
        \centering
        \includegraphics[width=\columnwidth, clip, trim = 0cm 1cm 0cm 0cm]{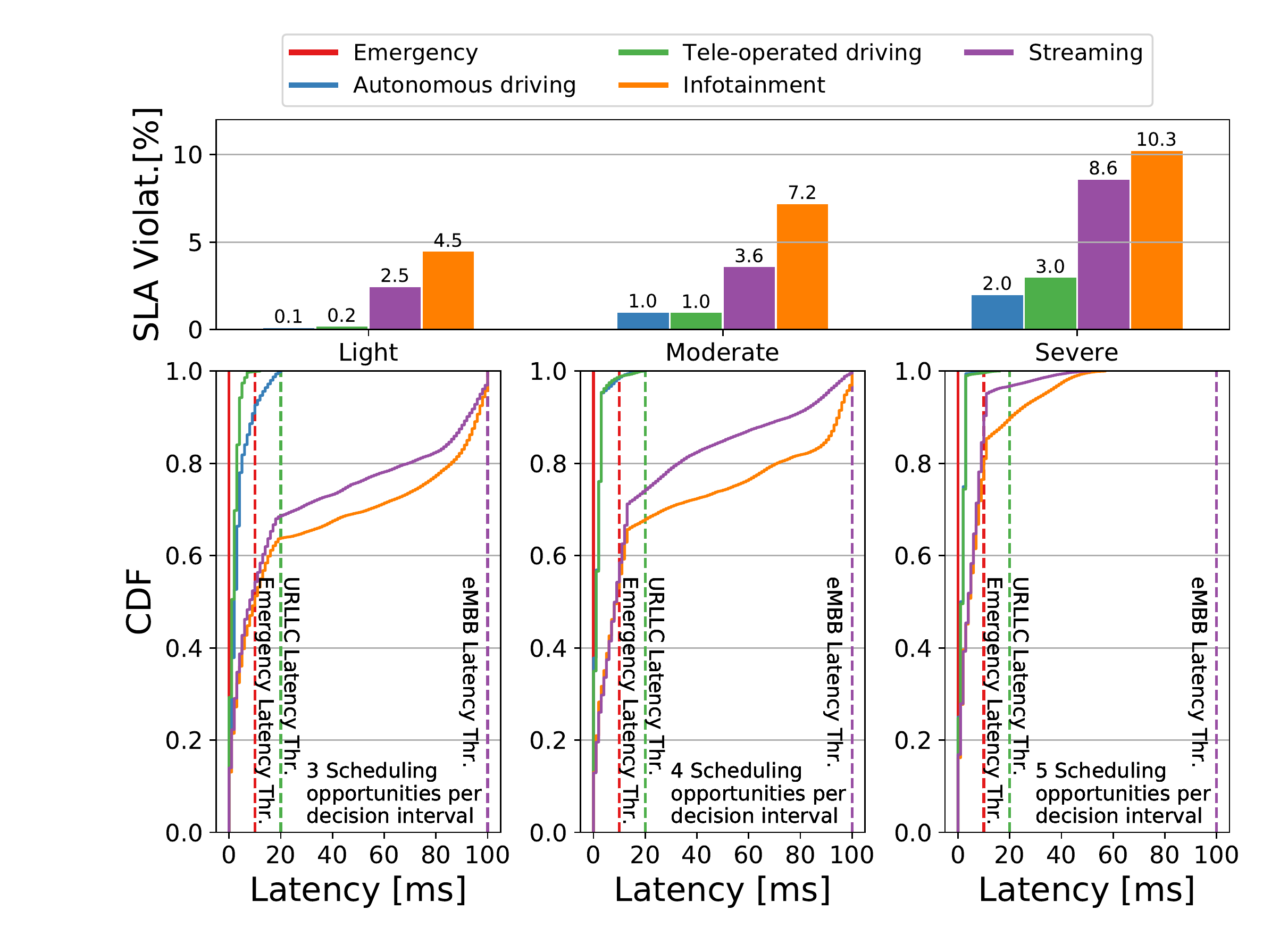}
        \caption{\small Resource orchestration at system level.}
        \label{fig:system_simulations}
    \end{subfigure}%
    ~
    \begin{subfigure}[t]{.5\textwidth}
        \centering
        \includegraphics[width=\columnwidth, clip, trim = 0cm 0cm 0cm 0cm]{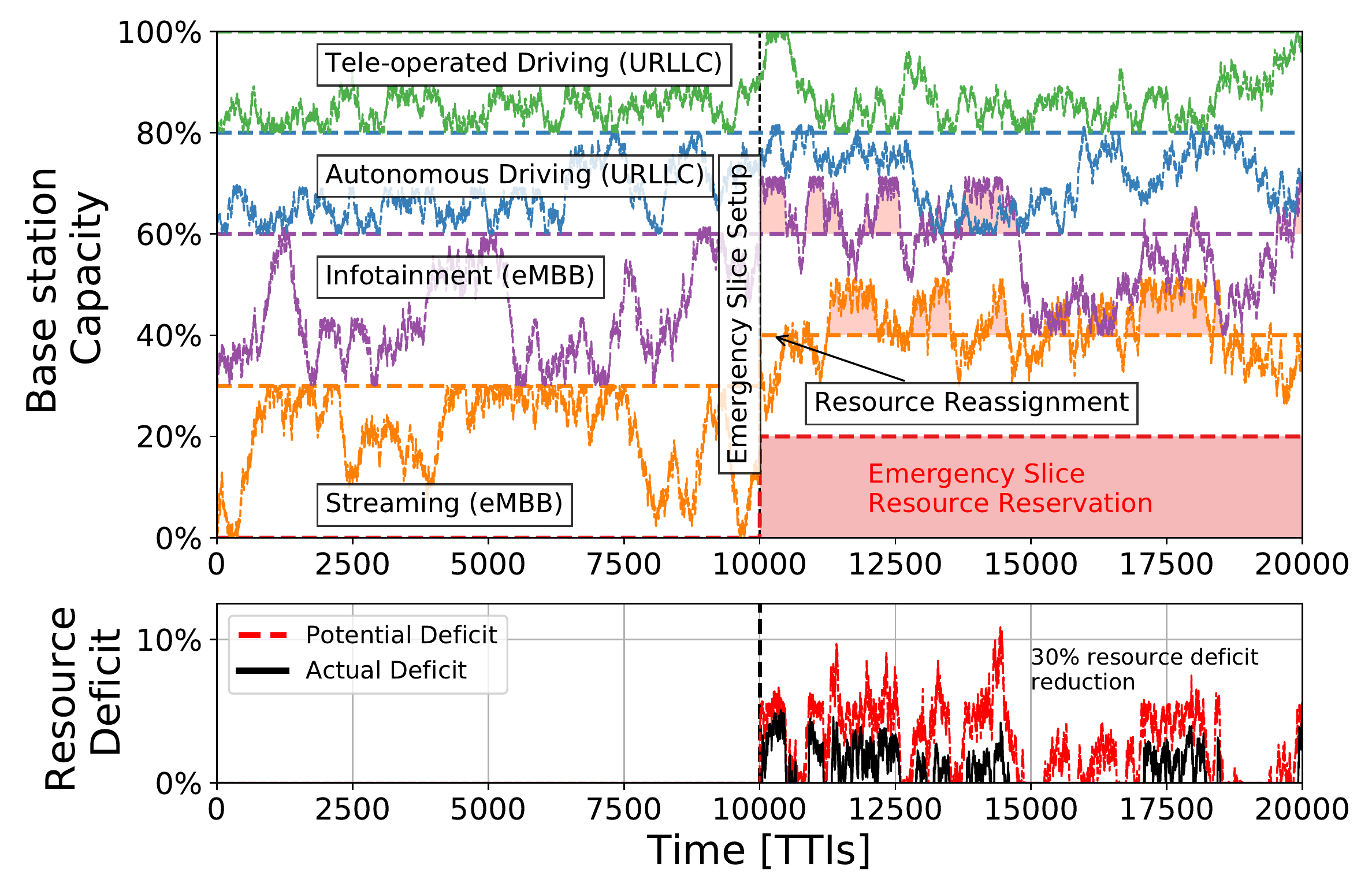}
        \caption{\small Resource orchestration at base station level.}
        \label{fig:tti_simulations}
    \end{subfigure}%
    \vspace{3mm}
    \caption{\small Performance evaluation of a $4$-slices scenario when an ENS is installed.}
    \label{fig:overall_results}
\end{figure*}

Intuitively, the impact of an ENS installation over the radio access network depends, among the others, by the networking capacity deployed in the area of the road event, instantaneous traffic demands, slice requirements, and user density. In Fig.~\ref{fig:system_simulations} (top) we provide a quantitative measure in case of different road event types and corresponding duration. We consider channel conditions as perceived by the base station during the event, thereby limiting the base station capacity (due to lower assigned MCS) when compared to the ideal scenario.
The results of each category are averaged over $10$ road event occurrences taken from our real dataset. It can be noticed that eMBB slices unveil worst performance (w.r.t. URLLC) due to their higher throughput requirements and lower slice priority (i.e., higher delay tolerance). However, an advanced slice orchestration solution may alleviate this issue. 

Fig.~\ref{fig:system_simulations} (bottom) focuses on the \textit{Moderate} road events evaluating the impact of different slice types (delay tolerance parameters $\lambda_s$) with a fixed decision interval $t$ duration of $100$ ms~\cite{URLLC_V2X}. 
Specifically, $\pi$-Orchestrator may suggest different solutions with corresponding slice configurations that translate into variable slice scheduling opportunities per decision interval. Therefore, we evaluate the Cumulative Distribution Functions (CDFs) of traffic latency considering different scheduling opportunities.
Despite the resource deficit introduced by the ENS, the average slice latency improves when the number of scheduling opportunities increases. 
Fig.~\ref{fig:tti_simulations} depicts the temporal evolution of the overall effects caused by the ENS slice setup at transmission time interval (TTI) level. 
All traffic traces, MCS and channel quality values are generated with a millisecond granularity according to their temporal evolution and statistical distributions resulting from our monitoring dataset. The initial scenario accounts for a fixed resource allocation scheme (without any ENS running) which allows to satisfy all the different slice requirements, as expected in normal traffic conditions thanks to the adoption of calibrated admission and control algorithms, e.g.~\cite{Slicing2018CoNEXT}. In particular, among the $4$ slices populating the system, eMBB slices have a fixed quota of $30$\% of available resources while URLLCs have $20$\%, respectively. After $10$ seconds, an ENS is instantiated. Its higher priority implies the execution of the $\pi$-Orchestrator to minimize service disruption. From the upper plot, it can be noticed how our solution reduces the quota of resources assigned to non-delay sensitive services which, unavoidably, suffer from a resource deficit (highlighted in red).
However, in emergency scenarios the RAN slice scheduler may assign unused reserved resources to active slices with traffic pending to be served, further reducing the resource deficit and, in turn, the SLA violation of certain slices. In the lower plot, we depict the overall traffic potentially violating the SLAs with a red dashed line, whereas the traffic actually exceeding it with a black continuous line. Results show how scheduling relaxation would help in reducing the SLA violations up to $30\%$. Clearly, this result assumes that none of the resources allocated for ENSs can be consumed by other slices, even in absence of emergency traffic.


\section{Related Work}
\label{sect:related}

The ever-increasing vehicular traffic fosters the need to deeply understand the complex relationship that regulates mobility patterns which, in turn, affects mobile network operational conditions~\cite{App_Layer_Prediction,Flow_Prediction_Spatio_Temporal}. From the one side, this effort requires constant monitoring of the communication infrastructure. From the other side, the highly heterogeneous set of monitoring KPIs demands for advanced solutions to automate the early detection of anomalous conditions.
The authors of~\cite{Automated_Diagnosis_2008} initially addressed this scenario proposing a Bayesian network working on a set of discrete metrics collected from an operational UMTS infrastructure. Similarly, the work of~\cite{BigData_Timeseries} focused on a large-scale cellular network scenario, where traffic traces are modeled into regular and random components. Their decomposition approach well suits predictable patterns, but fails in highly variable scenarios.
More recently, state-of-the-art solutions start combining monitoring traces and heterogeneous contextual information to improve the effectiveness of model predictions and decisions. In~\cite{DeepCogInfocom19, AztecInfocom20} the authors leverage spatio-temporal characteristics of mobile traces to predict resource utilization in the context of network slicing. 

Most of the works in the literature address the urban environment and aims to mitigate the cause of traffic congestions by predicting traffic flows and offloading strategies to alternative paths~\cite{LTE_Connectivity}, while less studies targeted the highway scenario. In~\cite{Mining_Spatio_Temporal} the authors present a spatio-temporal analysis of highways travel patterns exploiting per-vehicle data records collected by a centralized authority in China. Given the fine granularity of per-vehicle information, there is lack of a discussion about anomaly detection from a mobile network perspective.
Additionally, note that the majority of the related works presented above tackle the vehicular traffic prediction by means of active systems, e.g., GPS location exchange, or networks of sensors deployed along the road, which clearly ease the prediction task but also increases the volume of information generated along the process. To the best of our knowledge, this is the first work considering the prediction of road events and their impact on the mobile network infrastructure exclusively accounting for \emph{aggregated} and \emph{passive} RAN monitoring samples.


\section{Conclusions}
\label{sect:conclusions}
In this paper, we presented \name{}, a deep learning-based solution for the analysis and detection of road events over the highway. The model accounts for two complementary stages $i$) an autoencoder-based stage to identify anomalous patterns in the temporal evolution of operational mobile network data, and $ii$) a 3D-CNN-based stage to overcome simple threshold detection schemes and automatically learn the relationship that links multiple metric-specific anomalies to road event occurrences. 
The output of \name{} may be used in a multitude of emergency scenarios. We focused on the design and validation of a slice orchestration solution dealing with the setup of Emergency Network Slice (ENS) in V2X scenarios. Considering real mobile traffic distributions from a major highway in Italy, our results show that the information provided by \name{} can significantly reduce, up to $30\%$, the impact of Emergency Network Slice setup, also decreasing the probability of service disruption on other running network slices.


\bibliographystyle{IEEEtran}
\bibliography{main}
\end{document}